\documentclass[10pt,journal,compsoc]{IEEEtran}
\ifCLASSOPTIONcompsoc
  \usepackage[nocompress]{cite}
\else
  \usepackage{cite}
\fi
\usepackage{amsmath}
\usepackage{amssymb}
\usepackage{graphicx}
\usepackage{url}
\usepackage{color}
\usepackage{booktabs}
\usepackage{hyperref}
\usepackage{subfigure}

\newcommand{\remove}[1]{ }

\begin{document}

\title{Machine Learning for the Geosciences: Challenges and Opportunities}

\author{ Anuj Karpatne, Imme Ebert-Uphoff, Sai Ravela, Hassan Ali Babaie, and Vipin Kumar 
\IEEEcompsocitemizethanks{\IEEEcompsocthanksitem A. Karpatne is with the University of Minnesota.
\protect\\
Email: karpa009@umn.edu
\IEEEcompsocthanksitem I. Ebert-Uphoff is with the Colorado State University.
\protect\\
Email: iebert@colostate.edu
\IEEEcompsocthanksitem S. Ravela is with the Massachusetts Insititue of Technology.
\protect\\
Email: ravela@mit.edu
\IEEEcompsocthanksitem H. Babaie is with the Georgia State University.
\protect\\
Email: hbabaie@gsu.edu
\IEEEcompsocthanksitem V. Kumar is with the University of Minnesota.
\protect\\
Email: kumar001@umn.edu
}
}

\IEEEtitleabstractindextext{%
\begin{abstract}
%\begin{abstract} 
%\small\baselineskip=9pt 
% \comment{[Will be revised and expanded. Feel free to skip.]}
Geosciences is a field of great societal relevance that requires solutions to several urgent problems facing our humanity and the planet. As geosciences enters the era of big data, machine learning (ML)--- that has been widely successful in commercial domains---offers immense potential to contribute to problems in geosciences.
However, problems in geosciences have several unique challenges that are seldom found in traditional applications, requiring novel problem formulations and methodologies in machine learning. This article introduces researchers in the machine learning (ML) community to these challenges offered by geoscience problems and the opportunities that exist for advancing both machine learning and geosciences. We first highlight typical sources of geoscience data and describe their properties that make it challenging to use traditional machine learning techniques. We then describe some of the common categories of geoscience problems where machine learning can play a role, and discuss some of the existing efforts and promising directions for methodological development in machine learning. We conclude by discussing some of the emerging research themes in machine learning that are applicable across all problems in the geosciences, and the importance of a deep collaboration between machine learning and geosciences for synergistic advancements in both disciplines.

% properties of %\Imme{geoscience processes and}
% geoscience data that make machine learning challenging, such as their spatio-temporal structure, high dimensionality, heterogeneity in space and time, multi-scale/multi-resolution properties, small sample size, paucity of ground truth, noise, and uncertainty.  
% This is followed by a discussion of 
% key geoscience tasks that offer opportunities 
% for machine learning research, 
% and a discussion of 
% recent advances and directions
% in machine learning that have high potential for these tasks.

% \comment{This abstract is getting a little short - needs some updating?}
% \Anuj{Yes. But we can do it after we send it out to other authors.}
%Finally, we present our vision for geoscience-guided machine learning.
%This is followed by a discussion of resulting challenges and %opportunities of machine learning problems.  Overall, the rich %variety and complexity of challenges in geoscience problems %motivate novel frontiers of machine learning research. 
%This article was motivated by and builds on discussions at the 2015 NSF-funded {\it Workshop on Intelligent and Information Systems for Geosciences} (IS-GEO).

%\end{abstract}

\end{abstract}

% Note that keywords are not normally used for peerreview papers.
}

% make the title area
\maketitle

% To allow for easy dual compilation without having to reenter the
% abstract/keywords data, the \IEEEtitleabstractindextext text will
% not be used in maketitle, but will appear (i.e., to be "transported")
% here as \IEEEdisplaynontitleabstractindextext when the compsoc 
% or transmag modes are not selected <OR> if conference mode is selected 
% - because all conference papers position the abstract like regular
% papers do.
\IEEEdisplaynontitleabstractindextext
% \IEEEdisplaynontitleabstractindextext has no effect when using
% compsoc or transmag under a non-conference mode.

% For peer review papers, you can put extra information on the cover
% page as needed:
% \ifCLASSOPTIONpeerreview
% \begin{center} \bfseries EDICS Category: 3-BBND \end{center}
% \fi
%
% For peerreview papers, this IEEEtran command inserts a page break and
% creates the second title. It will be ignored for other modes.
\IEEEpeerreviewmaketitle

%\IEEEraisesectionheading{\section{Introduction}\label{sec:introduction}}
% Computer Society journal (but not conference!) papers do something unusual
% with the very first section heading (almost always called "Introduction").
% They place it ABOVE the main text! IEEEtran.cls does not automatically do
% this for you, but you can achieve this effect with the provided
% \IEEEraisesectionheading{} command. Note the need to keep any \label that
% is to refer to the section immediately after \section in the above as
% \IEEEraisesectionheading puts \section within a raised box.

% The very first letter is a 2 line initial drop letter followed
% by the rest of the first word in caps (small caps for compsoc).
% 
% form to use if the first word consists of a single letter:
% \IEEEPARstart{A}{demo} file is ....
% 
% form to use if you need the single drop letter followed by
% normal text (unknown if ever used by the IEEE):
% \IEEEPARstart{A}{}demo file is ....
% 
% Some journals put the first two words in caps:
% \IEEEPARstart{T}{his demo} file is ....
% 
% Here we have the typical use of a "T" for an initial drop letter
% and "HIS" in caps to complete the first word.
%\IEEEPARstart{G}{eoscience} data arise

% {\color{blue}
% [From Imme:  Switched to TKDE style file, but conversion is not complete yet.
% Abtract and keywords should appear above, but right now they do not.  Not sure why.]}

% {\color{blue}  [Imme says: How about simplifying the title to "Machine Learning for the Geosciences: Challenges and Opportunities" ?]}

%%%%%%%%%%%%%%%%%%%%%%%%%%%%%%%
\section{Introduction}

Momentous challenges facing our society require solutions to problems that are geophysical in nature \cite{kates2001sustainability,press2008earth,reid2010earth,intergovernmental2014climate}, such as predicting impacts of climate change, measuring air pollution, predicting increased risks to infrastructures by disasters such as hurricanes, modeling future availability and consumption of water, food, and mineral resources, and identifying factors responsible for earthquake, landslide, flood, and volcanic eruption.  
The study of such problems is at the confluence of several disciplines such as physics, geology, hydrology, chemistry, biology, ecology, and anthropology that aspire to understand the Earth system and its various interacting components, collectively referred to as the field of geosciences.

As the deluge of big data continues to impact practically every commercial and scientific domain, geosciences has also witnessed a major revolution from being a data-poor field to a data-rich field. This has been possible with the advent of better sensing technologies (e.g., remote sensing satellites and deep sea drilling vessels), improvements in computational resources for running large-scale simulations of Earth system models, and Internet-based democratization of data that has enabled the collection, storage, and processing of data on crowd-sourced and distributed environments such as cloud platforms.
Most geoscience data sets are publicly available and do not suffer from privacy issues that have hindered adoption of data science methodologies in areas such as health-care and cyber-security. The growing availability of big geoscience data offers immense potential for machine learning (ML)--- that has revolutionized almost all aspects of our living (e.g., commerce, transportation, and entertainment)---to significantly contribute to geoscience problems of great societal relevance.

Given the variety of disciplines participating in geoscience research and the diverse nature of questions being investigated, the analysis of geoscience data has several unique aspects that are strikingly different from standard data science problems encountered in commercial domains. For example, geoscience phenomena are governed by physical laws and principles and involve objects and relationships that often have amorphous boundaries and complex latent variables.  Challenges introduced by these characteristics motivate the development of new problem formulations and methodologies in machine learning that may be broadly applicable to problems even outside the scope of geosciences.

Thus, there is a great opportunity for machine learning researchers to closely collaborate with geoscientists and cross-fertilize ideas across disciplines for advancing the frontiers of machine learning as well as geosciences.
There are several communities working on this emerging field of inter-disciplinary collaboration at the intersection of  geosciences and machine learning. These include, but are not limited to,
Climate Informatics: a community of researchers conducting annual workshops to bridge problems in climate science with methods from statistics, machine learning, and data mining \cite{ci}; Climate Change Expeditions: a multi-institution multi-disciplinary collaboration funded by the National Science Foundation (NSF) Expeditions in Computing grant on ``Understanding Climate Change: A data-driven Approach'' \cite{expedition}; and
ESSI: a focus group of the American Geophysical Union (AGU) on Earth \& Space Sciences Informatics \cite{essi}.
More recently, NSF has funded a research coordination network on Intelligent Systems for Geosciences (IS-GEO) \cite{is-geo}, with the intent of forging stronger connections between the two communities.
Furthermore, a number of leading conferences in machine learning and data mining such as Knowledge Discovery and Data Mining (KDD), IEEE International Conferene on Data Mining (ICDM), SIAM International Conference on Data Mining (SDM), and Neural Information Processing Systems (NIPS) have included workshops or tutorials on topics related to geosciences. The role of big data in geosciences has also been recognized in recent perspective articles (e.g., \cite{faghmous2014big,monteleoni2013climate}) and special issues of journals and magazines (e.g., \cite{faghmous2015computing}).

The purpose of this article is to introduce researchers in the machine learning (ML) community to the opportunities and challenges offered by geoscience problems. The remainder of this article is organized as follows.
Section 2 provides an overview of the types and origins of geoscience data. 
Section 3 describes the challenges for machine learning arising from both the underlying geoscience processes and their data collection.  Section 4 outlines important geoscience problems where machine learning can yield major advances. Section 5 discusses two cross-cutting themes of research in machine learning that are generally applicable across all areas of geoscience.  Section 6 provides concluding remarks by briefly discussing the best practices for collaboration between machine learning researchers and geoscientists.

%%%%%%%%%%%%%%%%%%%%
\section{{Sources} 
of Geoscience Data}

The Earth and its major interacting components (e.g., lithosphere, biosphere, hydrosphere, and atmosphere) are complex dynamic systems \cite{Stillings:2012,Carbone:2016} in which the states of the system perpetually keep changing in space and time, in order to create a balance of mass and energy.  The elements of the Earth system (e.g., layers in oceans, ions in air, minerals and grains in rock, and land covers on the ground) interact with each other through complex and dynamic geoscience processes (e.g., rain falling on Earth's surface and nourishing the biomass, sediments depositing on river banks and changing river course, and magma erupting on sea floor and forming islands).
% The elements of the Earth system participate in a number of dynamic processes such as atmospheric flow, crystallization of lava, and weathering or deformation of rocks.  
% These processes, involving non-linear and multi-variate interactions, keep changing the states of the system in space and time, in order to create a balance in mass and energy. 
% In addition to interacting with elements in the same subsystem, the elements of every subsystem also interact with components of other external subsystems across the system boundaries.
% Due to the complex and varied nature of Earth science processes, geoscience data sets are available in diverse forms and characteristics, e.g., from Earth observing satellites~\cite{S13}, ground-based sensor measurements~\cite{S10}, cooperative autonomous observing systems~\cite{S6} or outputs of physics-based model simulations \cite{S8,S12,karpatne2013earth,karpatne2015guide}.

Data about these Earth system components and geoscience processes can generally be obtained from two broad categories of data sources: (a) observational data collected via sensors in space, in the sea, or on the land, and (b) simulation data from physics-based models of the Earth system. We briefly describe both these categories of gesocience data sources in the following. A detailed review of Earth science data sets and their properties can be found in \cite{karpatne2015guide}.

\subsection{Geoscience Observations}
Information about the Earth system is collected via different acquisition methods at varying scales of space and time and for a variety of geoscience objectives. For example, there is a nexus of Earth observing satellites in space that are tasked to monitor a number of geoscience variables such as surface temperature, humidity, optical reflectance, and chemical compositions of the atmosphere. There is a growing body of space research organizations ranging from public agencies such as the National Aeronautics and Space Administration (NASA), European Space Agency (ESA), and Japan Aerospace Exploration Agency (JAXA) to private companies such as SpaceX that are together contributing to the huge volume and variety of remote sensing data about our Earth, many of which are publicly available (e.g., see \cite{lpdaac}). Remote sensing data provides a global picture of the history of geoscience variables at fine spatial scales (1km to 10m, and less) and at regular time intervals (monthly to daily) for long periods, sometimes starting from the 1970s (e.g., Landsat archives \cite{landsat}). For targeted studies in specific geographic regions of interest, geoscience observations can also be collected using sensors on-board flying devices such as unmanned aerial vehicles (drones) or airplanes, 
{e.g., to detect and classify sources of methane (a powerful greenhouse gas) being emitted into the atmosphere \cite{frankenberg2016airborne}}.

Another major source of geoscience observations is the collection of \emph{in-situ} sensors placed on ground (e.g., weather stations) or moving in the atmosphere (e.g., weather balloons) or the ocean (e.g, ships and ocean buoys). Sensor-based observations of geoscience processes are generally available over non-uniform grids in space and at irregular intervals of time, sometimes even over moving bodies such as balloons, ships, or buoys. They constitute some of the most reliable and direct sources of information about the Earth's weather and climate systems and are actively maintained by public agencies such as the National Oceanic and Atmospheric Administration (NOAA) \cite{ncdc}. Sensor-based measurements from rain and river gauges are also central for understanding hydrological processes such as surface water discharge \cite{grdc}. Land-based seismic sensors, Global Positioning System (GPS)-enabled devices, and other geophysical instruments also continuously measure the Earth's geological structure and processes \cite{earthscope}.  In addition, we also have proxy measurements such as paleoclimatic records that are sparsely available at a select few locations but go back several thousands of years.

Given the huge variety in the characteristics of data for different geoscience processes, it is important to identify the type and properties of a given geoscience data set to make utmost use of relevant  data analytics methodologies. For example, remote sensing data sets, that are commonly available as rasters over regularly-spaced grid cells in space and time, can be represented as geo-registered images over individual time points or as time series data at individual spatial locations. On the other hand, sensor measurements from ships and ocean buoys can be represented as point reference data (also termed as geostatistical data in the spatial statistics literature) of continous spatio-temporal fields. Indeed, it is possible to convert one data type to another and across different spatial and temporal resolutions using simple interpolation methods or more advanced methods based on physical understanding such as reanalysis techniques \cite{kalnay1996ncep}.

% Knowledge of the data type can help in using the right methodologies. For example, raster data from satellites. point reference data from buoys. interconversions among data possible. Data is spatio-temporal. See STDM Survey.

\subsection{Earth System Model Simulations}
A unique aspect of geoscience processes is that the relationships among variables
or the evolution of states of the system are deeply grounded in physical laws and principles, discovered by the scientific community over multiple centuries of systematic research. For example, the motion of water in the lithosphere, or of air in the atmosphere, is governed by principles of fluid dynamics such as the Navier--Stokes equation. Although such physics-based equations can sometimes be solved in closed form for small-scale experiments, most often it is difficult to obtain their exact solutions for complex real-world systems encountered in the geosceiences. However, the underlying physical principles can still be used to \emph{simulate} the evolution of the states of the Earth system using numerical models referred to as physics-based models. Such models are the standard workhorse for studying a majority of geoscience processes where the state of the dynamical system can be time-stepped back in the past or forward in the future using inputs such as initial and boundary conditions or values of internal parameters in physical equations. Physics-based models generate large volumes of simulation data of different components of the Earth system, which can be used in data-driven analyses. They are developed and maintained by a number of centers constituting of diverse groups of researchers around the world. For example, the World Climate Reserach Programme (WCRP) develops and distributes simulations of General Circulation Models (GCM) of climate variables such as sea surface temperature and pressure under the Coupled Model Intercomparison Project (CMIP) \cite{cmip}. Simulations of terrestrial processes related to the lithosphere and biosphere are produced by the Community Land Model (CLM) \cite{clm}, developed by a number of international agencies collaborating with the National Center for Atmospheric Research (NCAR).

\section{Geoscience Challenges}
\label{data_properties_sec}

There are several characteristics of geoscience {applications} that limit the usefulness of traditional machine learning algorithms for knowledge discovery.
% in geoscience applications.  
% \Imme{
% Many of these characteristics were  enumerated in the 2015 IS-GEO workshop report \cite{IS-GEO:2015}, which motivated this section where we explore these characteristics in depth.}
Firstly, there are some inherent challenges arising from the nature of geoscience processes. For example, geoscience objects generally have amorphous boundaries in space and time that are not as crisply defined as objects in other domains, such as users on a social networking website, or products in a retail store. Geoscience phenomena also have spatio-temporal structure, are highly multi-variate, follow non-linear relationships (e.g., chaotic), show non-stationary characteristics, and often involve rare but interesting events. Secondly, apart from the inherent challenges of geoscience processes, the procedures used for collecting geoscience observations introduce more challenges for machine learning. This includes the presence of data at multiple resolutions of space and time, with varying degrees of noise, incompleteness, and uncertainties. Thirdly, for supervised learning approaches, there are additional challenges due to the small sample size (e.g., small number of historical years with adequate records) and lack of gold-standard ground truth in geoscience applications. In the following, we describe these three categories of geoscience challenges, namely (a) inherent challenges of geoscience processes, (b) geoscience data collection challenges, and (c) paucity of samples and ground truth, in detail.

% There are several characteristics of geoscience data %(Figure \ref{overview_plot}) 
% that limit the usefulness of traditional machine learning algorithms for knowledge discovery in geoscience applications.  First, there are some inherent challenges arising from the nature of geoscience processes, e.g., spatio-temporal nature of phenomena and presence of non-stationary relationships. Second, the approaches used for collecting geoscience data introduces more challenges for machine learning such as presence of multiple resolutions of data with varying degrees of noise, incompleteness, and uncertainties. Third, for supervised learning approaches, there are additional challenges due to the small sample size and lack of ground truth in geoscience applications. In the following, we describe these three categories of geoscience challenges in detail.

% Machine learning challenges arising due to these overarching properties are discussed in Section 3.1. There are also several challenges that arise due to the observation methodologies used for collecting geoscience data. This includes, for example, issues related to the complexity and quality of geoscience observations (discussed in Section 3.2) and paucity of samples and ground truth (discussed in Section 3.3).

%%%%%%%%%%%%%%%%%%%%%%%%%%%%%%%%%%%
\subsection{Inherent Challenges of Geoscience Processes}
\label{sec:challenge_overarching}
\remove{
The primary challenges for machine learning emerge from nonlinearity, dimensionality, and uncertainty. This characterization provided by Ravela and McLaughlin \cite{S11} follows others that have identified dimensionality \cite{S14} 
and nonlinearity \cite{S16}. 
The equations that describe geophysical processes can be nonlinear, even chaotic, spanning a broad range of spatial and temporal scales involving multivariate multiphase state variables and the absence of a dynamical “slow manifold”. 
Resolution and scales in computational models must be filtered to simulate such processes tractably. Questions of closure representation of fine-scale processes are thus left open. As regional and smaller scales become of interest, the increasing dimensionality of the numerical models compounds the problem.
}
% \\[0.3cm]
%
%
%\begin{enumerate}  
% Decided NOT to use enumerate here, because the indentation takes away too much space.
%
{\bf Property 1: Objects with Amorphous Boundaries}\\
Geoscience objects include waves, flows, and coherent structures in all phases of matter.
% and their dynamical transport and transmission mechanisms. 
Hence, the form, structure, and patterns of geoscience objects that can exist at multiple scales in continuous spatio-temporal fields are much more complex than those found in “discrete spaces” that machine learning algorithms typically deal with, such as items in market basket data. For example, eddies, storms and hurricanes dynamically deform in complicated ways from a purely object-oriented perspective. New techniques to consider both the pattern and dynamical information of coherent objects and their uncertainties are being developed \cite{S4,S7}, but new methods for capturing other features of geoscience objects, e.g., fluid segmentation and fluid feature characterization, are needed.
\\[0.3cm]
{\bf Property 2: Spatiotemporal Structure}\\
%
% \comment{Imme says: The following paragraph has a lot of repetitions.  E.g.\ teleconnections are mentioned early on, then discussed more later. Is that on purpose?}
% Geoscience data are naturally distributed across space and time with characteristic heavy-tails~\cite{S3,S12} emerging that contain not just local space-time correlations but significant long-range teleconnections.
Since almost every geoscience phenomena occurs in the realm of space and time, geoscience observations are generally auto-correlated in both space and time when observed at appropriate spatial and temporal resolutions.
For example, a location that is covered by a certain land cover label (e.g., forest, shrubland, urban) is generally surrounded by locations that have similar land cover labels. Land cover labels are also consistent along time, i.e., the label at a certain time is related to the labels in its immediate temporal vicinity. Furthermore, if the land cover at a certain location changes (e.g., from forest to croplands), the change generally persists for some temporal duration instead of switching back and forth.

Although spatio-temporal autocorrelation dictates stronger connectivity among nearby observations in space and time, geoscience processes can also show long-range spatial dependencies. For example, a commonly studied phenomenon in climate science is teleconnections \cite{wallace1981teleconnections,kawale2013graph}, where two distant regions in the world show strongly coupled activity in climate variables such as temperature or pressure. Geoscience processes can also show long-memory characteristics in time, e.g., the effect of climate indices such as the El Ni\~{n}o Southern Oscillation (ENSO) and Atlantic Multidecadal Oscillation (AMO) on global floods, droughts, and forest fires \cite{siegert2001increased,ward2014strong}.

The inherent spatio-temporal structure of geoscience data has several implications on machine learning methods. This is because many of the widely used machine learning methods are founded on the assumption that observed variables are independent and identically distributed ($i.i.d$). However, this assumption is routinely violated in geoscience problems, where variables are structurally related to each other in the context of space and time, unless there is a discontinuity, such as a fault, across which autocorrelation ceases to persist. Cognizance of the spatio-temporal autocorrelation in geoscience data collected in continuous media is crucial for the effective modeling of geophysical phenomena.
\\[0.3cm]
{\bf Property 3: High Dimensionality}\\
The Earth system is incredibly complex, with a huge number of potential variables, which may all impact each other, and thus many of which may have to be considered simultaneously~\cite{S14}.  For example, the robust and complete detection of land cover changes, such as forest fires, requires the analysis of multiple remote sensing variables, such as vegetation indices and thermal anomaly signals. Capturing the effects of these multiple variables at fine resolutions of space and time renders geoscience data inherently high dimensional, where the number of dimensions can easily reach orders of millions.

As an example, in order to study processes occurring on the Earth's surface, even a relatively coarse resolution data set (e.g. at 2.5$^o$ spatial resolution) may easily result in more than 10,000 spatial grid points, where every grid point has multiple observations in time. Furthermore, geoscience phenomena are not limited to the Earth's surface, but extend beneath the Earth's surface (e.g., in the study of groundwater, faults, or petroleum) and across multiple layers in the atmosphere or the mantle, additionally increasing the dimensionality of the data in 3D spatial resolutions. Hence, there is a need to scale existing machine learning methods to handle tens of thousands, or millions, of dimensions for global analysis of geoscience phenomena.
\\[0.3cm]
{\bf Property 4: Heterogeneity in Space and Time}\\
An interesting characteristic of geoscience processes is their degree of variability in space and time, leading to a rich heterogeneity in geoscience data across space and time. For example, due to the presence of varying geographies, vegetation types, rock formations, and climatic conditions in different regions of the Earth, the characteristics of geoscience variables vary significantly from one location to the other.  Furthermore, the Earth system is not stationary in time and goes through many cycles, ranging from seasonal and decadal cycles to long-term geological changes (e.g., glaciation, polarity reversals) and even climate change phenomena, that impact all local processes. This heterogeneity of geoscience processes makes it difficult to study the joint distribution of geoscience variables across all points in space and time. Hence, it is difficult to train machine learning models that have good performance across all regions in space and across all time-steps. Instead, there is a need to build local or regional models, each corresponding to a homogeneous group of observations. 
% This requires accounting for the “context” of observation in geoscience data sets as input parameters in machine learning algorithms. 
\\[0.3cm]
{\bf Property 5: Interest in Rare Phenomena}\\
In a number of geoscience problems, we are interested in studying objects, processes, and events that occur infrequently in space and time but have major impacts on our society and the Earth's ecosystem. For example, extreme weather events such as cyclones, flash floods, and heat waves can result in huge losses of human life and property, thus making it vital to monitor them for adaptation and mitigation requirements. These processes may relate to emergent (or anomalous) states of the Earth system, or other features of complex systems such as anomalous state trajectories and basins of attractions \cite{babaie2017ontology}.
As another example, detecting rare changes in the Earth's biosphere such as deforestation, insect damage, and forest fires can be helpful in assessing the impact of human actions and informing decisions to promote ecosystem sustainability. Identifying such rare classes of changes and events from geoscience data and characterizing their behavior is challenging. This is because we often have an inadequate number of data samples from the rare class due to the skew (imbalance) between the classes, making their modeling and characterization difficult. 
% Secondly, evaluating the performance of a model in the presence of rare classes is challenging as standard evaluation metrics, such as accuracy and false positive rate, can be quite insensitive to false positives when the skew among the classes is large, thus leading to spurious insights.

%%%%%%%%%%%%%%%%%%%%%%%%%%%%%%%%%%%
\subsection{Geoscience Data Collection Challenges}

% Apart from the inherent challenges in studying geoscience processes, there are additional challenges introduced by the procedures used for collecting geoscience observations. 
% We discuss two of these challenges in this category, namely the availability of geoscience observations at varying scales and resolutions, and the poor quality of geoscience data, e.g., due to noise, incompleteness and uncertainties in the data.
% % We highlight one important property that causes geoscience observational data to be particularly complex.
% \\[0.3cm]
%
{\bf Property 6: Multi-resolution Data}\\
Geoscience data sets are often available via different sources (e.g., satellite sensors, in-situ measurements and model-based simulations) and at varying spatial and temporal resolutions. These data sets may exhibit varying characteristics, such as sampling rate, accuracy, and uncertainty. For example, in-situ sensors, such as buoys in the ocean and hydrological and weather measuring stations, are often irregularly spaced.  As another example, collecting high-resolution data of ecosystem processes, such as forest fires, may require using aerial imageries from planes flying over the region of interest, which may need to be combined with coarser resolution satellite imageries available at frequent time intervals. The analysis of multi-resolution geoscience data sets can help us characterize processes that occur at varying scales of space and time. For example, processes such as plate tectonics and gravity occur at a global scale, while local processes include volcanism, earthquakes, and landslides.  To handle multi-resolution data, a common approach is to build a bridge between data sets at disparate scales (e.g., using interpolation techniques), so that they can be represented at the same resolution. We also need to develop algorithms that can identify patterns at multiple resolutions without upsampling all the data sets to the highest resolution. 
\\[0.3cm]
{\bf Property 7: Noise, Incompleteness, and Uncertainty in Data}\\
Many geoscience data sets (e.g., those collected by Earth observing satellite sensors) are plagued with noise and missing values. For example, sensors may temporarily fail due to malfunctions or severe weather conditions, resulting in missing data. Additionally, changes in measuring equipment, e.g., replacing a faulty sensor or switching from one satellite generation to the next, may change the interpretation of sensor values over time, making it difficult to deploy a consistent methodology of analysis across different time periods. In many geoscience applications, the signal of interest can be small in magnitude compared to the magnitude of noise. 
%For example, the atmosphere is a highly chaotic system which causes a low signal-to-noise ratio in the observed data. 
%
%Hence, geoscience measurements naturally have a high degree of uncertainty, i.e. we can rarely be certain that the observations represent the exact state of the desired variable. 
%
Furthermore, many sensor properties can increase noise, such as sensor interference, e.g., in the case of remotely sensed land surface data, where atmospheric (clouds and other aerosols) and surface (snow and ice) interference are constantly encountered. 

%Also, while model outputs are very high dimensional, uncertainty propagation at higher resolutions is challenging.

Many geoscience variables cannot even be measured directly, but can only be inferred from other observations or model simulations. For example, one can use airborne imaging spectrometers to detect sources of methane (e.g., pipeline leaks), an important greenhouse gas. These instruments fly overhead surveys and map the ground-reflected sunlight arriving at the sensor. Methane plumes can then be identified from excess sunlight absorption \cite{Thompson1}.  But to determine the leak rate (flux) and the resulting greenhouse gas impact, one must also know how fast the excess methane mass is dispersing.  This requires considering the influence of air transport, which in turn requires steady-state physical assumptions, morphology-based plume modeling, or direct in-situ measurements of the wind speed.
Even data generated from model outputs have uncertainties because of our imperfect knowledge of the initial and boundary conditions of the system or the parametric forms of approximations used in the model. 

% Because of the poor quality of geoscience data, machine learning researchers need to work in close collaboration with geoscience researchers to fully understand the context of the variables, necessary preprocessing steps, and appropriate interpretation and visualization of results, \Imme{as discussed in Section \ref{collaboration_sec}}. The poor quality of data also motivates the need for developing machine learning methods that are robust to the presence of a high degree of noise and missing values.  

%%%%%%%%%%%%%%%%%%%%%%%%%%%%%%%%%%%
\subsection{Paucity of Samples and Ground Truth}

\label{sec:paucity}
% Another challenge in working with geoscience problems is that we often have very few samples (e.g. small number of historical years with adequate records) or gold-standard ground truth~\cite{S12}. Challenges arising due to these data limiting characteristics are discussed in the following. 
% % This section discusses why data sets in geosciences often contain 
% % very few samples, or at least very few labeled samples.
% \\[0.3cm]
%
%%
{\bf Property 8: Small sample size}\\
The number of samples in geoscience data sets is often limited in both space and time. Factors that limit sample size include history of data collection and the nature of phenomenon being measured. For example, most satellite products are only available since the 1970s, and when monthly (yearly) processes are considered, this means that less than 600 (50) samples are available. Furthermore, there are many events in geosciences that are important to monitor but occur very infrequently, thus resulting in small sample sizes. For example, a majority of land cover changes, landslide, tsunami, and forest fire are rare events, and only occur for short temporal durations mostly over small spatial regions.  With less than 80 years of reliable sensor-based data, only a few dozens of rare events are available as training data.

The limited spatial and temporal resolution of some geoscience variables is also limited by the nature of observation methodology. For example, paleo-climate data are derived from coral, lake sediments (varves), tree rings, and deep ice core samples, which are only available at a few places around the Earth. Similarly, early records of precipitation only exist in areas covered by land. 

% \comment{Imme says: Moved the following paragraph here from Property 9. I believe it belongs here.  Do you agree?}
% \Imme{
% Furthermore, there are many events in geoscience data that are important to monitor but occur very infrequently. For example, a majority of land cover changes, landslide, tsunami, and forest fire are rare events, and only occur for short temporal durations mostly over small spatial regions.  Another example is the occurrence of El Ni\~{n}o/La Ni\~{n}a or hurricane events, which can occur only on a yearly basis. With less than 80 years of related sensor data available, and since events do not occur every year, only dozens of events are available as training data.}

This is in contrast to commercial applications involving Internet-scale data, e.g., text mining or object recognition, where large volumes of labeled or unlabeled data have been one of the major factors behind the success of machine learning methodologies such as deep learning. The limited number of samples in geoscience applications along with the large number of physical variables result in problems that are under-constrained in nature, requiring novel machine learning advances for their robust analyses.
\\[0.3cm]
{\bf Property 9: Paucity of Ground Truth}\\
Even though many geoscience applications involve large amounts of data, e.g., global observations of ecosystem variables at high spatial and temporal resolutions using Earth observing satellites, a common feature of geoscience problems is the paucity of labeled samples with gold-standard ground truth. This is because high-quality measurements of several geoscience variables can only be taken by expensive apparatus such as low-flying airplanes, or tedious and time-consuming operations such as field-based surveys, which severely limit the collection of ground truth samples. Other geoscience processes (e.g., subsurface flow of water) do not have ground truth at all, since, due to complexity of the system, the exact state of the system is never fully known.  

The paucity of representative training samples can result in poor performance for many machine learning methods, either due to underfitting where the model is too simple, or due to overfitting where the model is overly complex relative to the dimensionality of features and the limited number of training samples. Hence, there is a need to develop machine learning methods that can learn parsimonious models even in the paucity of labeled data. Another possibility is to construct synthetic data sets through simulations \cite{ED:2017} or perturbation that can be used for training~\cite{S13}, to make the most of the few observations.  %Providing such data sets is one of the goals of the IS-GEO Working group on Case Studies 
%(see \url{http://organicdatacuration.org/is-geo/}
%\url{index.php/Geoscience_Case_Studies}).  
% Lastly, the development of methods that fuse traditional ML methods and domain knowledge is of interest here as well~\cite{S1,S18}.

%\\[0.3cm]
%
%Uncertainty pervades our investigation of the Earth System in a variety of ways~\cite{S1,S2,S5,S7,S11}. 
%
%
%{\bf Property 9:	 Uncertainty in Data}\\
%
%

% These data Uncertain initial conditions and, for that matter, boundary conditions, forcing (inputs), and fine-scale representations (parameterizations) all point to an inherently stochastic approach to prediction and parameterization, and place strong demands on estimation and quantification of uncertainty in predictions. For example, at regional scales, the uncertainties become non-Gaussian \cite{S2,S5,S4,S7}. 
%Direct observations of the Earth system are sparse and noisy. 
%Further, although the advent of satellites has improved observational methods, 
% the entire depth of the ocean and the atmosphere are under-represented. 
% Measurements contain significant outliers, and missing data is standard.  
% They are sampled unevenly in space and time, and often the analysis methods become obsolete within a fraction of the timescales of interest.
%
%
%{\bf Property 9: Increased Uncertainty in Inferred Observations}\\
%In many cases, the physical variables of interest must be “retrieved” from the model outputs using %inversion procedures, which can introduce additional uncertainties due to improper inference. 

%%%%%%%%%%%%%%%%%%%%%%%%%%%%%%%%%%%
\section{Geoscience Problems and ML Directions}
%\label{Opportunities_sec}

Geoscientists constantly strive to develop better approaches for modeling the current state of the Earth system (e.g., how much methane is escaping into the atmosphere right now,  which parts of Earth are covered by what kind of biomass) and its evolution, as well as the connections within and between all of its subsystems (e.g., how does a warming ocean affects specific ecosystems).  
This is aimed at advancing our scientific understanding of geoscience processes. This can also help in providing actionable information (e.g., extreme weather warnings) or  informing policy decisions that directly impact our society (e.g., adapting to climate change and progressing towards sustainable lifestyles).  
The boundaries between these goals often blur in practice, e.g., an improved tornado model may simultaneously lead to a better science model as well as a more effective warning system.

Viewing from the lens of geosciences, many methods from machine learning are a natural fit for the problems encountered in geoscience applications. For example, classification and pattern recognition methods are useful for characterizing objects such as extreme weather events or swarms of foreshocks or aftershocks (tremors preceding or following an earthquake), estimating geoscience variables, and producing long-term forecasts of the state of the Earth system. As another example, approaches for mining relationships and causal attribution can provide insights into the inner workings of the Earth system and support policy making.  
% Since it is generally difficult to glean through the wealth of background information and identify the relevant geoscience problem to be investigated, understanding the different categories of geoscience problems is important. 
In the following, we briefly describe five broad categories of geoscience problems, and discuss promising machine learning directions and examples of some recent successes that are relevant for each problem.

\remove{
Claire: non-linear estimation
Jennifer estimation

Yan Liu: prediction
Amy: Prediction

}

%Numerous machine learning algorithms have some application in the geosciences, but most need to be adjusted to deal with the challenging data properties outlined in Section \ref{data_properties_sec}, and new algorithms need to be developed to fill the remaining gaps.  

%%%%%%%%%%%
%\subsection{Sample Geoscience Tasks}

% In other instances the science tasks are far removed from the applications, which makes it harder for the machine learning expert to spot those tasks.  
% Below we discuss a variety of such tasks, including both science and application tasks. 

%%%%%%%%%%%
\subsection{Characterizing Objects and Events}

Machine learning algorithms can help in characterizing objects and events in geosciences that are critical for understanding the Earth system. 
For example, we can analyze patterns in geoscience data sets to detect climate events such as  cyclogenesis and tornadogenesis, and discover their precursors for predicting them with long leads in time. Analyzing spatial and temporal patterns in geoscience data can also help in studying the formation and movement of climate objects such as weather fronts, atmospheric rivers, and ocean eddies, which are major drivers of vital geoscience processes such as the transfer of precipitation, energy, and nutrients in the atmosphere and ocean.
% This is particularly challenging, because of the amorphous spatial and temporal boundaries of these objects/events, and particularly important, because these events have large societal impact.

While traditional approaches for characterizing geoscience objects and events are primarily based on the use of hand-coded features (e.g., ad-hoc rules on size and shape constraints for finding ocean eddies \cite{chelton2007global}), machine learning algorithms can enable their automated detection from data with improved performance using pattern mining techniques.
However, in the presence of spatio-temporal objects with amorphous boundaries and their associated uncertainties \cite{S7}, there is a need to develop  pattern mining approaches that can account for the spatial and temporal properties of geoscience data  while characterizing objects and events. One such approach has been successfully used for finding spatio-temporal patterns in sea surface height data \cite{faghmous2013parameter,faghmous2014spatio}, resulting in the creation of a global catalogue of mesoscale ocean eddies \cite{faghmous2015daily}. Another approach for finding anomalous objects buried under the surface of the Earth (e.g., land mines) from radar images was explored in \cite{pe2017automated}, using unsupervised techniques that can work with mediums of varying properties. The use of topic models has also been explored for finding extreme events from climate time series data \cite{tang2015can}.

% In particular, the notion of spatio-temporal consistency \cite{faghmous2014spatio}, where objects are required to have consistent and significant signatures in space and time, has to be fully integrated in existing pattern mining algorithms.

%%%%%%%%%%%
\subsection{Estimating Geoscience Variables from Observations}

There is a great opportunity for machine learning methods to infer critical geoscience variables that are difficult to monitor directly, e.g., methane concentrations in air or groundwater seepage in soil, using information about other variables collected via satellites and ground-based sensors, or simulated using Earth system models. For example, supervised machine learning algorithms can be used to analyze remote sensing data and produce estimates of ecosystem variables such as forest cover, health of vegetation, water quality, and surface water availability, at fine spatial scales and at regular intervals of time. 
Such estimates of geoscience variables can help in informing management decisions and enabling scientific studies of changes occurring on the Earth's surface.

\begin{figure}[tbh]
\centering
\subfigure[Absolute resdiual errors of the baseline method.]{\label{global_map} \includegraphics[width=0.23\textwidth]{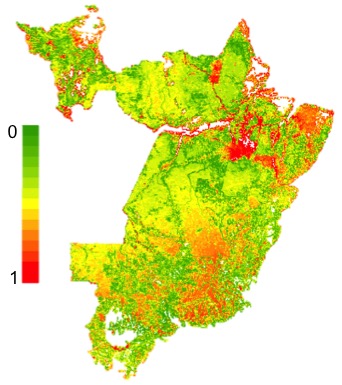}}
% ~ \quad
\subfigure[Absolute resdiual errors of the multi-task learning method presented in \cite{karpatnepredictive2014}.]{\label{mtl_map} \includegraphics[width=0.23\textwidth]{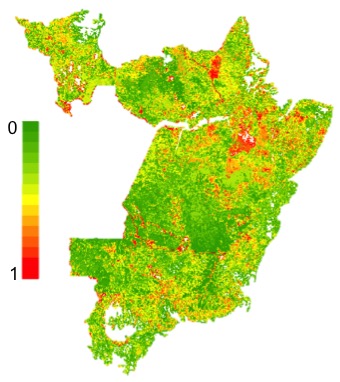}}
\caption{Performance improvement in estimation of forest cover in four states of Brazil using a multi-task learning method. Figure courtesy: Karpatne et al. \cite{Karpatne:2016}.}
\label{mtl_fig}
% \vspace{-0.25cm}
\end{figure}

A major challenge in the use of supervised learning approaches for estimating geoscience variables is the heterogeneity in the characteristics of variables across space and time. One way to address this challenge of heterogeneity is to explore multi-task learning frameworks \cite{baxter2000model,caruanamultitask}, where the learning of a model at every homogeneous partition of the data is considered as a separate task, and the models are shared across similar tasks to regularize their learning and avoid the problem of overfitting, especially when some tasks suffer from paucity of training samples. An example of a multi-task learning based approach for handling heterogeneity can be found in a recent work in \cite{karpatnepredictive2014}, where the learning of a forest cover model at every vegetation type (discovered by clustering vegetation time-series at locations) was treated as a separate task, and the similarity among vegetation types (extracted using hierarchical clustering techniques) was used to share the learning at related tasks. Figure \ref{mtl_fig} shows the improvement in prediction performance of forest cover in Brazil using a multi-task learning approach. A detailed review of promising machine learning advances such as multi-task learning, multi-view learning, and multi-instance learning for addressing the challenges in supervised monitoring of land cover changes from remote sensing data is presented in \cite{Karpatne:2016}.

To address the non-stationary nature of climate data, online learning algorithms have been developed to combine the ensemble outputs of expert predictors (climate models) and produce robust estimates of climate variables such as temperature \cite{monteleoni2011tracking,mcquade2012global}. In this line of work, weights over experts were updated in an adaptive way across space and time, to capture the right structure of non-stationarity in the data. This was shown to significantly outperform the the baseline technique used in climate science, which is the non-adaptive mean over experts (multi-model mean). Another approach for addressing non-stationarity  was explored in \cite{das2014non}, where a Bayesian mixture of models was learned for downscaling climate variables, where a different model was learned for every homogeneous cluster of locations in space. In a recent work, adaptive ensemble learning methods \cite{karpatne2015adaptive,karpatne2015ensemble}, in conjunction with physics-based label refinement techniques \cite{khandelwal2015post}, have been developed to address the challenge of heterogeneity and poor data quality for mapping the dynamics of surface water bodies using remote sensing data \cite{khandelwal2017approach}. This has enabled the creation of a global surface water monitoring system (publicly available at \cite{watermonitor}) that is able to discover a variety of changes in surface water such as shrinking lakes due to droughts, melting glacial lakes, migrating river courses, and constructions of new dams and reservoirs.

Another challenge in the supervised estimation of geoscience variables is the small sample size and paucity of ground-truth labels. Methods for handling the problem of high dimensions and small sample sizes have been explored in \cite{chatterjee2012sparse}, where sparsity-inducing regularizers such as sparse group Lasso were developed to model the domain characteristics of climate variables. To address the paucity of labels, novel learning frameworks such as semi-supervised learning, that leverages the structure in the unlabeled data for improving classification performance \cite{ssl-survey}, and active learning, where an expert annotator is actively involved in the process of model building  \cite{settles2010active}, have huge potential for improving the state-of-the-art in estimation problems encountered in gesocience applications \cite{vatsavai2005semi,tuia2009active}. In a recent line of work, attempts to build a machine learning model to predict forest fires in the tropics using remote sensing data led to a novel methodology for building predictive models for rare phenomena \cite{mithal2017rapt} that can be applied in any setting where it is not possible to get high quality labeled data even for a small set of samples, but poor quality labels (perhaps in the form of heuristics) are available for all samples. 

In addition to supervised learning approaches, given the plentiful availability of unlabeled data in geoscience applications such as remote sensing, there are several opportunities for unsupervised learning methods in estimating geoscience variables. For example, changes in time series of vegetation data, collected by satellite instruments on fixed time intervals at every spatial location on the Earth's surface, have been extensively studied using unsupervised learning approaches for mapping land cover changes such as deforestation, insect damage, farm conversions, and forest fires \cite{Verbesselt2010,mithal2011monitoring,chandola2011scalable}.

% There is also a great opportunity for intelligent sensing to adaptively gather data \cite{S6} and control the instrumentation of in-situ sensors to match the evolving dynamical regime of geoscience processes.
% See also XXX below.

%A related task that geoscientists are very interested in is {\it data assimilation}, namely the question of how to best integrate observed data into an existing simulation model to improve prediction accuracy.  This is by no means a trivial task, as geoscience variables are highly coupled in models and thus one cannot simply alter the value of one variable that is available from observation, e.g.\ temperature, without propagating the values of unobserved variables accordingly. 

%%%%%%%%%%%
\subsection{Long-term Forecasting of Geoscience Variables}
Predicting long-term trends of the state of the Earth system, e.g., forecasting geoscience variables ahead in time, can help in modeling future scenarios and devising early resource planning and adaptation policies. One approach for generating forecasts of geoscience variables is to run physics-based model simulations, which basically encode geoscience processes using state-based dynamical systems where the current state of the system is influenced by previous states and observations using physical laws and principles. From a machine learning perspective, this can be treated as a time-series regression problem where the future conditions of a geoscience variable has to be predicted based on present and past conditions. 
Some of the existing methods for time-series forecasting include exponential smoothing techniques \cite{gardner2006exponential}, autoregressive integrated moving average (ARIMA) models \cite{box1976time}, state-space models \cite{aoki2013state}, and probabilistic models such as hidden Markov models and Kalman filters \cite{rabiner1986introduction,harvey1990forecasting}. Machine learning methods for forecasting climate variables using the spatial and temporal structure of geoscience data have been explored in recent works such as \cite{liu2012sparse,mcgovern2015solar,gagne2015day,gagne2017storm}.

A key challenge in predicting the long-term trends of geoscience variables is to develop approaches that can represent and propagate prediction uncertainties, which is particularly difficult due to the high-dimensional and non-stationary nature of geoscience processes \cite{gel2004calibrated,gel2016catching}. In the climate scenario, there is limited long-term predictability at fine spatial scales necessary for implementing policy decisions. Some advances have been made in downscaling future projections to high spatial resolutions, e.g., using physics-based Markov Chain and Random Field models \cite{S12}, but much remains to be done. Further, the data is sparse, and the uncertainty distributions remain poorly sampled~\cite{S11,S1}. The heavy-tailed nature of extreme events such as cyclones and floods further exacerbates the challenges in producing their long-term forecasts. In a recent work \cite{liu2012sparse}, regression models based on extreme value theory have been developed to automatically discover sparse temporal dependencies and make predictions in multivariate extreme value time series. Other approaches for predicting extreme weather events such as abnormally high rainfall, floods, and tornadoes using climate data have also been explored in \cite{zhuang2016evaluation,wang2015hierarchical,yu2015tornado}. Effective prediction of geoscience variables can benefit from recent advances in machine learning such as transfer learning \cite{pan2010survey}, where the model trained on a present task (with sufficient number of training samples) is used to improve the prediction performance on a future task with limited number of training samples.

%%%%%%%%%%%
\subsection{Mining Relationships in Geoscience Data}
\label{relationships_sec}

An important problem in geoscience applications is to understand how different physical processes are related to each other, e.g., periodic changes in the sea surface temperature over eastern Pacific Ocean---also known as the El Ni\~{n}o-Southern Oscillation (ENSO)---and their impact on several terrestrial events such as floods, droughts, and forest fires \cite{siegert2001increased,ward2014strong}. Identifying such relationships from geoscience data can help us capture vital signs of the Earth system and advance our understanding of geoscience processes. A common class of relationships that is studied in the climate domain is \emph{teleconnections}, which are pairs of distant regions that are highly correlated in climate variables such as sea level pressure or temperature. One of the widely-studied category of teleconnections is dipoles \cite{kawale2013graph, kawale2011discovering}, which are pairs of regions with strong negative correlations (e.g., the ENSO phenomena). There is a huge potential in discovering such relationships using data-driven approaches, that can sift through vast volumes of observational and model-based geoscience data and discover interesting patterns corresponding to geoscience relationships.

One of the first attempts in discovering relationships from climate data is a seminal work by Steinbach et al. \cite{steinbach2003discovery}. In this work, graph-based representations of global climate data were constructed in which each node represents a location on the Earth and an edge represents the similarity (e.g., correlation) between the climate time series observed at a pair of locations. 
Dipoles and other higher-order relationships (e.g., tripoles involving triplets of  regions) could then be discovered from climate graphs using clustering and pattern mining approaches. Another family of approaches for mining relationships in climate science is based on representing climate graphs as complex networks \cite{tsonis2004architecture}. This includes approaches for examining the structure of the climate system \cite{donges2009complex}, studying hurricane activity \cite{elsner2009visibility}, and finding communities in climate networks \cite{steinhaeuser2010exploration,steinhaeuser2011complex}.

Formidable challenges arise in the problem of relationship mining due to the enormous search space of candidate relationships, and the need to simultaneously extract spatio-temporal objects, their relationships, and their dynamics, from noisy and incomplete geoscience data. Hence, there is a need for novel approaches that  can directly discover the relationships as well as the interacting objects 
\cite{agrawal2017tripoles,kawale2013graph}. For example, recent work on the development of such approaches have led to the discovery of previously unknown climate phenomena \cite{liess2017teleconnection,lu2016exploring,liess2014different}.

%%%%%%%%%%%
\subsection{Causal Discovery and Causal Attribution}

\begin{figure}
\centering
\includegraphics[width=7.0cm,angle=0]
{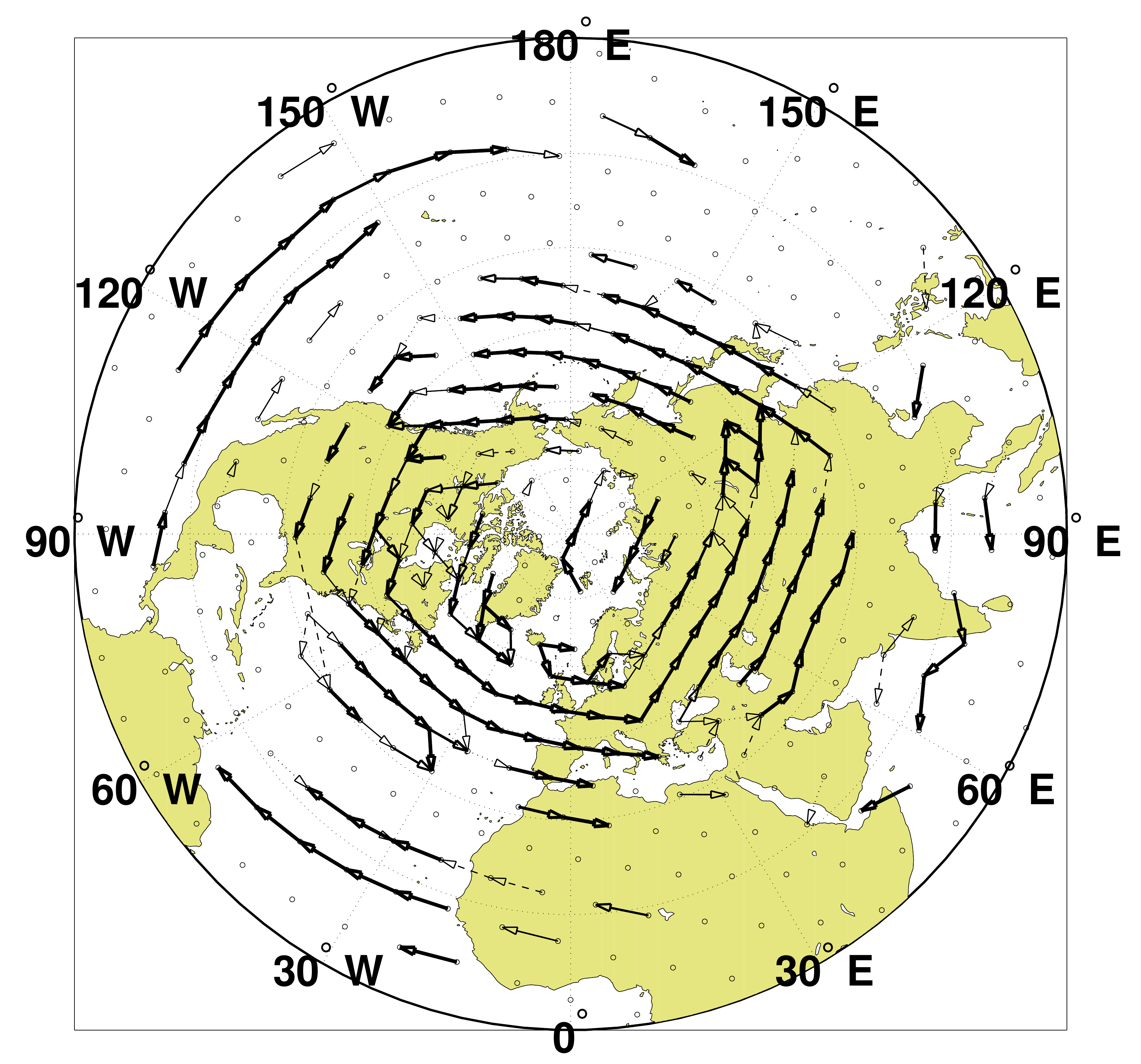}
\caption{Network plot for Northern hemisphere generated from daily geopotential height data using constraint-based structure learning of graphical models. The resulting arrows represent the pathways of storm tracks, see \cite{ED_ICMLA}. 
\label{Fekete_800_fig}}
\end{figure}

Discovering cause-effect relationships is an important task in the geosciences, closely related to the task of learning relationships in geoscience data, discussed in Subsection \ref{relationships_sec}. 
The two primary frameworks for analyzing cause-effect relationships are based on the concept of {\it Granger causality} \cite{granger1969investigating}, which defines causality in terms of {\it predictability}, and on the concept of {\it Pearl causality} \cite{pearl1991theory}, which defines causality in terms of {\it changes resulting from intervention}.  
Currently, the most common tool for causality analysis in the geosciences is bivariate Granger analysis, followed by multi-variate Granger analysis using vector autoregression (VAR) models \cite{lozano2009spatial}, but the latter is still not commonly used. Pearl's framework based on probabilistic graphical models has only rarely been used in the geosciences to date \cite{ED:2012,ED:2017,causal_attribution}.
The fact that such multi-variate causality tools, which have yielded tremendous breakthroughs in biology and medicine over the past decade, are still not commonly used in the geosciences, is in stark contrast to the huge potential these methods have for tackling numerous geoscience problems. These range from variable selection for estimation and prediction tasks to identifying causal pathways of interactions around the globe, (see Figure \ref{Fekete_800_fig}), and causal attribution \cite{lozano2009spatial,causal_attribution}. The latter is discussed in more detail below. 

%%%%%%%%%%%
%\subsection{Causal Attribution and Decision Making}

Many components of the Earth system are affected by human actions, thus introducing the need for integrating policy actions in the modeling approaches. The outputs produced by geoscience models can help inform policy  and decision making. 
The science of causal attribution is an essential tool for decision making that helps scientists determine the causes of events.
The framework of causal calculus \cite{Pearl:causality_book} provides a concise terminology for causal attribution of extreme weather and climate events \cite{causal_attribution}.
Methods based on Graphical Granger models have also been proposed \cite{lozano2009spatial},
but neither framework has been widely used.
Of great interest is the development of decision methodology with uncertain prediction probabilities, producing ambiguous risk with poorly resolved tails representing the most interesting extreme, rare, and transient events produced by models. The application of reinforcement learning and other stochastic dynamic programming approaches that can solve decision problems with ambiguous risk \cite{S3} are promising directions that need to be pursued.

\section{Cross-cutting Research Themes}

% \Imme{The preceeding sections outlined numerous ways in which} machine learning can play a significant role in advancing the frontiers of our knowledge in several geoscience tasks. 
In this section, we discuss two emerging themes of machine learning research that are generally applicable across all problems of geosciences. This includes deep learning and the paradigm of theory-guided data science, as described below.

%%%%%%%%%%%%%%
\subsection{Deep Learning}

Artificial neural networks have had a long and winding history spanning more than six decades of research, starting from humble origins with the perceptron algorithm in 1960s to present-day ``deep'' architectures consisting of several layers of hidden nodes, dubbed as \emph{deep learning} \cite{goodfellow2016deep}. The power of deep learning can be attributed to its use of a deep hierarchy of latent features (learned at hidden nodes), where complex features are represented as compositions of simpler features. This, in conjunction with the availability of big labeled data sets \cite{deng2009imagenet}, computational advancements for training large networks, and algorithmic improvements for back-propagating errors across deep layers of hidden nodes have revolutionized several areas of machine learning, such as supervised, semi-supervised, and reinforcement learning. Deep learning has resulted in major success stories in a wide range of commercial applications, such as computer vision, speech recognition, and natural language translation. 

Given the ability of deep learning methods to extract relevant features automatically from the data, they have a huge potential in geoscience problems where it is difficult to build hand-coded features for objects, events, and relationships from complex geoscience data. Owing to the space-time nature of geoscience data, geoscience problems share some similarity with problems in computer vision and speech recognition, where deep learning has achieved major accomplishments using frameworks such as convolutional neural networks (CNN) and recurrent neural networks (RNN), respectively. For example, if a CNN can learn to recognize objects such as cats in images, it could also be used to recognize objects and events such as tornadoes, hurricanes, and atmospheric rivers, which show structural features (e.g., sinkholes) in geoscience data. Indeed, the use of CNNs for detecting extreme weather events from climate model simulations has recently been explored in \cite{liu2016application,racah2016semi}. Similarly, RNN based frameworks such as long-short-term-memory (LSTM) models have been explored for mapping plantations in Southeast Asia from remote sensing data, using spatial as well as temporal properties of the dynamics of plantation conversions \cite{jia2017incremental,jia2017predict,jia2016learning}. Such frameworks are able to extract the right length of memory needed for making predictions in time, and thus can be useful for forecasting geoscience variables with appropriate lead times. Deep learning based frameworks have also been explored for downscaling outputs of Earth system models and generating climate change projections at local scales \cite{vandal2017deepsd}, and classifying objects such as trees and buildings in high-resolution satellite images \cite{basu2015deepsat}. These efforts highlight the promise of using deep learning to obtain similar accomplishments in geosciences as in the commercial arena, by incorporating the characteristics of geoscience processes (e.g., spatio-temporal structure) in deep learning frameworks. While the availability of large volumes of labeled data have been one of the major factors behind the success of deep learning in commercial domains, a key challenge in geoscience problems is the paucity of labeled samples, thus limiting the effectiveness of traditional deep learning methods. There is thus a need to develop novel deep learning frameworks for geoscience problems, that can overcome the paucity of labeled data, {for example,} by using domain-specific information of physical processes.

%%%%%%%%%%%%%%
\subsection{Theory-Guided Data Science}

% One of the basic goals in geosciences is to represent relationships between variables, e.g., the effect of carbon emission on the global dynamics of climate variables such as temperature. This is achieved with the help of physics-based models that either involve solving closed-form equations or running computational simulations of dynamical systems. 
% Despite the usefulness of physics-based models in explaining several geoscience processes, they suffer from certain knowledge gaps in problems where our current physical understanding is limited, mainly because of the inherent complexity of the processes, e.g., the flow and transport of groundwater beneath the Earth's surface, or the physics behind the creation and movement of clouds. In such settings, physics-based models are often forced to make many parametric approximations about the physical processes, which not only leads to poor predictive performance, but also renders the model difficult to comprehend and analyze. 

% An alternate possibility is to automatically learn relationships between geoscience variables using data science models, which have found great success in commercial applications, e.g., in vision and speech. However, solely relying on the knowledge contained in the limited number of data samples can lead to physically inconsistent models that are not only uninterpretable, but also suffer from poor predictive performance due to the phenomen\Imme{on} of overfitting. 
Given the complexity of problems in geoscience applications and the limitations of current methodological frameworks in geosciences (e.g., see recent debate papers in hydrology \cite{gupta2014debates,lall2014debates, mcdonnell2014debates}), neither a data-only, nor a physics-only, approach can be considered sufficient for knowledge discovery. Instead, there is an opportunity to pursue an alternate paradigm of research that explores the continuum between physics (or theory)-based models and data science methods by deeply integrating scientific knowledge in data science methodologies, termed as the paradigm of theory-guided data science \cite{tgds}. For example, scientific consistency can be weaved in the learning objectives of predictive learning algorithms, such that the learned models are not only less complex and show low training errors, but are also consistent with existing scientific knowledge. This can help in pruning large spaces of models that are inconsistent with our physical understanding, thus reducing the variance without likely affecting the bias. Hence, by anchoring machine learning frameworks with scientific knowledge, the learned models can stand a better chance against overfitting, especially when training data are scarce. For example, a recent work explored the use of physics-guided loss functions for tracking objects in sequences of images \cite{stewart2017label}, where elementary knowledge of laws of motion was solely used for constraining outputs and learning models, without the help of training labels.
Another motivation for learning physically consistent models and solutions is that they can be easily understood by domain scientists and ingested in existing knowledge bases, thus translating to scientific advancements.

The paradigm of theory-guided data science is beginning to be pursued in several scientific disciplines ranging from material science to hydrology, turbulence modeling, and biomedicine. A recent paper \cite{tgds} builds the foundation of this paradigm and illustrates several ways of blending scientific knowledge with data science models, using emerging applications from diverse domains. There is a great opportunity for exploring similar lines of research in geoscience applications, where machine learning methods can play a major role in accelerating knowledge discovery by automatically learning patterns and models from the data, but without ignoring the wealth of knowledge accumulated in physics-based model representations of geoscience processes  \cite{S1}.
This can complement existing efforts in the geosciences on integrating data in physics-based models, e.g., in model calibration, where parametric forms of approximations used in models are learned from the data by solving inverse problems, or in data assimilation, where the sequence of state transitions of the system are informed by measurements of observed variables wherever available \cite{evensen2009introduction}.

\section{Conclusions}
\label{Conclusions_sec}

The Earth System is a place of great scientific interest that impacts every aspect of life on this planet and beyond.
The survey of challenges, problems, and promising machine learning directions provided 
in this article is clearly not exhaustive, but it 
illustrates the great emerging possibilities
of future machine learning research in this important area.
% Because of the poor quality of geoscience data, machine learning researchers need to work in close collaboration with geoscience researchers to fully understand the context of the variables, necessary preprocessing steps, and appropriate interpretation and visualization of results

Successful application of machine learning techniques in the geosciences is generally driven by a science question arising in the geosciences, and the best recipe for success tends to be for a machine learning researcher to collaborate very closely with a geoscientist {during all phases} of research.
That is because the geoscientists are in a better position to understand which science question is novel and important, {which variables and} data set to use to answer that question, the strengths and weaknesses inherent in the data collection process that yielded the data set, and which pre-processing steps to apply, such as smoothing or removing seasonal cycles.  Likewise, the machine learning researchers are better placed to decide which data analysis methods are available and appropriate for the data, the strengths and weaknesses of those methods, and what they can realistically achieve.  
% Thus the researchers from these two communities need to {work as a team and make all major decisions together}, starting from the selection of the science question to be investigated, to the selection of the data set and  the analysis method, and the interpretation of results. 
Interpretability is also an important end goal in geosciences because if we can understand the basic reasoning behind the patterns, models, or relationships extracted from the data, they can be used as building blocks in scientific knowledge discovery. Hence, choosing methods that are inherently transparent 
are generally preferred in most geoscience applications.
Further, the end results of a study 
need to be translated into geoscience language so that it can be related back to the original science questions. Hence, frequent communication between the researchers avoids long detours and ensures that the outcome of the analysis is indeed rewarding for both machine learning researchers and geoscientists \cite{ED_EOS}.

\ifCLASSOPTIONcaptionsoff
  \newpage
\fi

\section*{Acknowledgements}
The authors of this paper are supported by inter-disciplinary projects at the interface of machine learning and geoscience, including the NSF Expeditions in Computing grant on ``Understanding Climate Change: A Data-driven Approach'' (Award \#1029711), the NSF-funded 2015 IS-GEO workshop (Award \#1533930), and subsequent Research Collaboration Network (EarthCube RCN IS-GEO: Intelligent Systems Research to Support Geosciences, Award \#1632211). 
% \comment{[Add acknowledgments to your funding resources relevant for this paper.]}
The vision outlined in this paper has been greatly influenced by the collaborative works in these projects.
In particular, the description of geoscience challenges in Section \ref{data_properties_sec} has been movtiated by the initial discussions at the 2015 IS-GEO workshop (see workshop report \cite{IS-GEO:2015}).

% We thank the NSF for funding the  and the subsequent .  \Imme{This work was also supported by XXX}

\remove{
The 2015 NSF-funded workshop on Intelligent and Information Systems for Geosciences (IS-GEO) was the brain child of the two workshop chairs, Yolanda Gil and Suzanne A.\  Pierce. 
We thank all participants of the workshop for their contributions to the discussions, namely Hassan Babaie, Arindam Banerjee, Kirk Borne, Gary Bust, Michelle Cheatham, Imme Ebert-Uphoff, Carla Gomes, Mary Hill, John Horel, Leslie Hsu, Jim Kinter, Craig Knoblock, David Krum, Vipin Kumar, Pierre Lermusiaux, Yan Liu, Deborah McGuinness, Chris North, Victor Pankratius, Shanan Peters, Beth Plale, Allen Pope, Sai Ravela, Juan Restrepo, Aaron Ridley, Hanan Samet, Shashi Shekhar, Katie Skinner, Padhraic Smyth, Basil Tikoff, Lynn Yarmey, and Jia Zhang. 
We thank the NSF for funding the 2015 IS-GEO workshop (Award \#1533930) and the subsequent Research Collaboration Network (EarthCube RCN IS-GEO: Intelligent Systems Research to Support Geosciences, Award \#1632211).
To join this research collaboration network, visit \url{is-geo.org}.
}

\bibliographystyle{IEEEtran}
\bibliography{ref}

\remove{

}

% biography section
% 
% If you have an EPS/PDF photo (graphicx package needed) extra braces are
% needed around the contents of the optional argument to biography to prevent
% the LaTeX parser from getting confused when it sees the complicated
% \includegraphics command within an optional argument. (You could create
% your own custom macro containing the \includegraphics command to make things
% simpler here.)
%\begin{IEEEbiography}[{\includegraphics[width=1in,height=1.25in,clip,keepaspectratio]{mshell}}]{Michael Shell}
% or if you just want to reserve a space for a photo:

% \newpage

\begin{IEEEbiography}[{\includegraphics[height=1in]{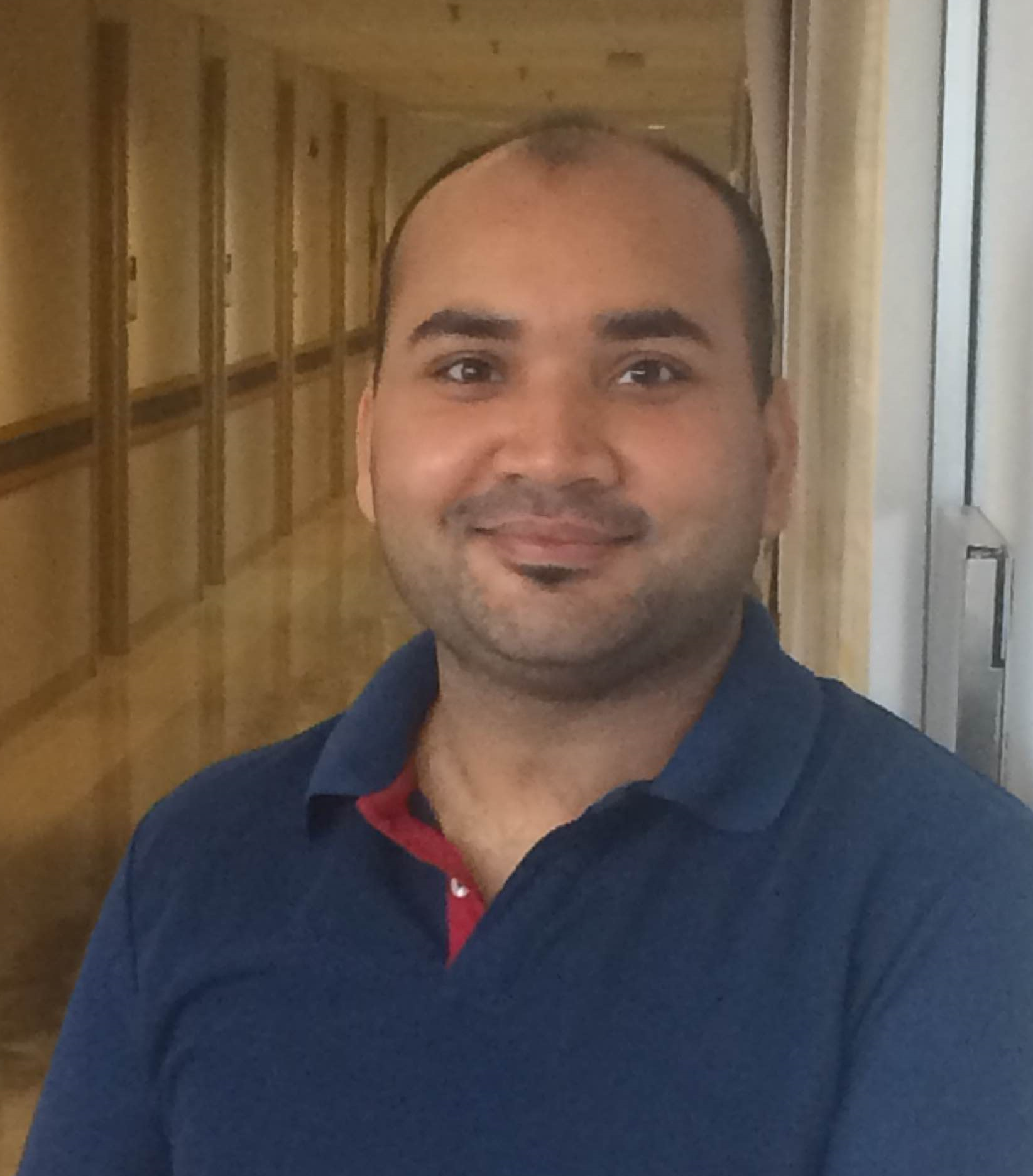}}]{Anuj Karpatne}
Anuj Karpatne is a PhD candidate in the Department of Computer Science and Engineering (CSE) at University of Minnesota (UMN). 
Karpatne works in the area of data mining with applications in scientific problems related to the environment. 
Karpatne received his B.Tech-M.Tech degree in Mathematics \& Computing from Indian Institute of Technology (IIT) Delhi.
\end{IEEEbiography}

\begin{IEEEbiography}[{\includegraphics[height=1in]{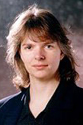}}]{Imme Ebert-Uphoff}
Imme Ebert-Uphoff is a Research Faculty member in the Department of Electrical and Computer Engineering at Colorado State University.  Her research interests include causal discovery and other machine learning methods, applied to geoscience applications.  She received her Ph.D. in Mechanical Engineering from the Johns Hopkins University (Baltimore, MD), and her M.S. and B.S. degrees in Mathematics from the University of Karlsruhe (Karlsruhe, Germany). 
\end{IEEEbiography}

\begin{IEEEbiography}[{\includegraphics[height=1in]{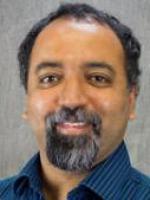}}]{Sai Ravela}
Sai Ravela directs the Earth Signals and Systems Group (ESSG) in the Earth, Atmospheric and Planetary Sciences at the Massachusetts Institute of Technology. His primary research interests are in dynamic data-driven stochastic systems  theory and machine intelligence methodology with application to Earth, Atmospheric and Planetary Sciences. Ravela received a PhD in Computer Science in 2003 from the University of Massachusetts at Amherst. 
\end{IEEEbiography}

\begin{IEEEbiography}[{\includegraphics[height=1in]{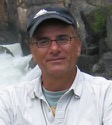}}]{Hassan Ali Babaie}
Hassan Ali Babaie is an Associate Professor at the Department of Geosciences, with joint appointment in the Computer Science Department, at Georgia State University. Babaie’s research interest includes geoinformatics, semantic web, representing the knowledge of structural geology, applying ontologies, and machine learning. He received his PhD specializing in Structural Geology from Northwestern University. 
\end{IEEEbiography}

\begin{IEEEbiography}[{\includegraphics[height=1in]{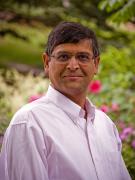}}]{Vipin Kumar}
Vipin Kumar is a Regents Professor and William Norris Chair in Large Scale Computing in the Department of CSE at UMN.
Kumar's research interests include data mining, high-performance computing, and their applications in climate/ecosystems and biomedical domains.
Kumar received his PhD in Computer Science from University of Maryland, M.E in Electrical Engineering from Philips International Institute Eindhoven, and B.E in Electronics \& Communication Engineering from IIT Roorkee.
\end{IEEEbiography}

\end{document}